\documentclass[11pt]{article}

\usepackage[final]{mtsummit25}
\usepackage{copyright}
\usepackage{times}
\usepackage{latexsym}
\usepackage[T1]{fontenc}
\usepackage[utf8]{inputenc}
\usepackage{microtype}
\usepackage{inconsolata}
\usepackage{graphicx}
\usepackage{booktabs}
\usepackage{tabularx}
\usepackage{fancyvrb}
\usepackage{fvextra}
\usepackage{fixltx2e}
\usepackage{nth}
\usepackage{dirtree}

\usepackage{lineno} % for line numbers
\nolinenumbers % disable line numbers globally

\title{Testing LLMs' Capabilities in Annotating Translations Based on an Error Typology Designed for LSP Translation: First Experiments with ChatGPT}

\author{
  Joachim Minder\textsuperscript{1, 2} \and 
  Guillaume Wisniewski\textsuperscript{2} \and
  Natalie Kübler\textsuperscript{1} \\
  \textsuperscript{1}Université Paris Cité, ALTAE, F-75013 Paris, France \\
  \textsuperscript{2}Université Paris Cité, CNRS, Laboratoire de linguistique formelle, F-75013 Paris, France
}

\begin{document}
\pagestyle{plain}

\maketitle
\thispagestyle{plain}

\newcommand{\err}[1]{\colorbox{orange!10}{#1}}

\begin{abstract}
This study investigates the capabilities of large language models (LLMs), specifically ChatGPT, in annotating MT outputs based on an error typology. In contrast to previous work focusing mainly on general language, we explore ChatGPT's ability to identify and categorise errors in specialised translations. By testing two different prompts and based on a customised error typology, we compare ChatGPT annotations with human expert evaluations of translations produced by DeepL and ChatGPT itself. The results show that, for translations generated by DeepL, recall and precision are quite high. However, the degree of accuracy in error categorisation depends on the prompt's specific features and its level of detail, ChatGPT performing very well with a detailed prompt. When evaluating its own translations, ChatGPT achieves significantly poorer results, revealing limitations with self-assessment. These results highlight both the potential and the limitations of LLMs for translation evaluation, particularly in specialised domains. Our experiments pave the way for future research on open-source LLMs, which could produce annotations of comparable or even higher quality. In the future, we also aim to test the practical effectiveness of this automated evaluation in the context of translation training, particularly by optimising the process of human evaluation by teachers and by exploring the impact of annotations by LLMs on students' post-editing and translation learning.
\end{abstract}

\section{Introduction}

As underlined by the famous quote attributed to  Yorick Wilk: “More has been written about MT evaluation than about MT itself”~\citep{king-etal-2003-femti}. Translation evaluation is an essential but highly challenging task. It relies primarily on two approaches. The first consists in assigning scores that reflect translation quality at different levels ---~be it a segment, a paragraph, a document, or a system. This type of metric is central to assessing machine translation performance. The second approach, more commonly used in the field of translation studies and translation training, although increasingly used in MT evaluation (see, e.g., \newcite{freitag-etal-2021-experts}), involves annotating translations by identifying errors (i.e.\ words that need to be corrected to improve the translation) and categorising them according to an error typology.

The high cost of human evaluation, whether in terms of time, technical expertise, effort, or financial resources, has driven researchers to explore ways to automate evaluation. While automatic metrics capable of approximating human judgments, with varying degrees of accuracy, have existed for quite some time and continue to improve \citep{marie-etal-2021-scientific}, automating error annotation remains a significantly more complex challenge. Until recently, it had attracted little attention and was even considered out of reach.

The launch of ChatGPT in 2022, and more generally the development of LLMs since the 2020s, opened up new possibilities for automating this second type of evaluation. LLMs, initially designed to produce fluid text in natural language, quickly drew the attention of the scientific community for their ability to perform complex tasks, for which they have not been explicitly programmed, taking advantage of their ability to model and manipulate language. Numerous experiments have indeed highlighted the possibility of using LLMs to solve tasks simply by prompting them, i.e.\ by explaining in natural language how to solve the task at hand. In recent years, pioneering works, reviewed in Section~\ref{sec:rw}, have emerged, highlighting the potential of LLMs, notably ChatGPT, to automate translation evaluation, reflecting a growing interest in integrating LLMs into tasks that, until now, have relied mainly on human intervention. Our work follows this dynamic by investigating the possibility of prompting an LLM (here ChatGPT) to annotate MT outputs by identifying and categorising errors. However, we explore this possibility in a new direction: our experiments stand out by focusing exclusively on specialised translation (LSP, language for specific purposes), in opposition to previous research, which primarily considers general language. Specialised translation has considerable economic implications, as it plays a crucial role in industries ranging from law and medicine to technology. Specialised translation also raises additional challenges, both for human translators/evaluators and for MT systems. These include accurate processing of specialised terminology, phraseology, and the management of complex patterns inherent to specialised texts.

We have carried out different annotation experiments using ChatGPT with different prompts and two different MT systems frequently used by professional and non-professional translators alike (DeepL or ChatGPT). Our goals with these experiments are:
\begin{itemize}
    \item To evaluate the effectiveness of our prompts with ChatGPT when annotating specialised translations (in this case, in the field of natural language processing);
    \item To analyse ChatGPT's performance in error identification and categorisation, based on an error typology designed for specialised translation evaluation;
    \item To compare ChatGPT's annotation performances with respect to the MT system being evaluated.
\end{itemize}
Ultimately, by shedding light on this issue, our work aims to contribute to the creation of hybrid tools, where artificial intelligence and human expertise complement each other to promote more effective learning and teaching, and more accessible self-evaluation or teacher evaluation.

The rest of this work is organised as follows. We will start by reviewing related works in Section~\ref{sec:rw}, before detailing the context and motivations of our work. We will then present our experimental setting in Section~\ref{sec:exp_method} and the results of the different experiments we have carried out in Section~\ref{sec:results}.

\section{Related Works \label{sec:rw}}

In recent years, the NLP and translation community has shown interest in exploiting LLMs as translators, with promising results. Many even imply that the future of MT is closely linked to LLMs and generative AI (cf.\ e.g.\ \newcite{lyu-etal-2024-paradigm,wang-etal-2023-document-level,he-2024-prompting,jiao2023chatgpt,siu2023ChatGPT}). If LLMs are able to produce translations, they should also be able to assess translations and distinguish between high and low-quality translations. This assumption was the starting point for a group of researchers in NLP and translation studies who, in early 2023, began exploring the ability of LLMs to evaluate translations. 

\paragraph*{Using LLMs to predict human judgements} One of the pioneering works in this field is that of \newcite{kocmi-etal-2023-findings}, who created the GEMBA metric in zero-shot mode\footnote{Zero-shot learning is a machine learning paradigm in which a model can make predictions by leveraging semantic knowledge, such as descriptions, rather than relying solely on labeled training examples.}, both with and without reference. Their primary goal was to compare the evaluations performed by 9 different GPT models with reference human annotations from WMT'22 \citep{kocmi-etal-2022-findings} and to observe the level of correlation between the 2 types of evaluation, both at the system and segment level. GPT-based evaluations were carried out with scoring (direct assessment and Scalar Quality Metric SQM) and classification (quality classes) tasks. Their experience on 3 high-resource language pairs shows that, with direct assessment, GEMBA with reference achieves state-of-the-art performance in comparison with other WMT'22 metrics. Without reference, i.e., for quality estimation tasks, GEMBA is the best metric. The best performance is achieved when using GPT-3.5 and higher models, especially GPT-4 \citep{openai2024gpt4technicalreport}.

\paragraph*{Using LLMs for Error Annotation} Later, \newcite{lu-etal-2024-error} went one step further by using ChatGPT to evaluate translation quality through error analysis (EA) prompting using WMT'20 data \citep{barrault-etal-2020-findings}, again with high-resource languages. Their goal is still to compare the evaluations of ChatGPT (made here in a few-shot and chain-of-thought (CoT) mode) with reference annotations:  they asked ChatGPT to identify minor and major errors based on the MQM typology\footnote{\url{https://themqm.org/}} and to score them in order to achieve state-of-the-art performance at both the system and segment levels. Their results show that their EA metric achieves state-of-the-art performance at the system level, but lags behind other metrics at segment level. However, they show that combining CoT and EA improves evaluation capabilities at the segment level, provided that the prompt includes examples (few-shot).

Another work is that of \newcite{fernandes-etal-2023-devil}, who created the AutoMQM metric with and without reference, aiming to identify and classify errors according to MQM (interpretable metric) and produce a quality score with the PaLM \citep{chowdhery24palm} and PaLM-2 models~\citep{anil23palm2}, but with both high-resource languages using WMT'22 data \citep{kocmi-etal-2022-findings} and low-resource languages using WMT'19 data \cite{ma-etal-2019-results}. They aim to show how fine-tuning on human annotation data boosts the performances of LLMs. Their results show that prompted AutoMQM achieves state-of-the-art performance at the system level, but that fine-tuning is necessary to boost performance at the segment level, especially without reference. They also show that adding in-context examples to prompts improves model performance. Experiments with low-resource languages show that LLMs are still underperforming.

With this growing interest in applying error analysis to LLM-based evaluation, \newcite{kocmi-federmann-2023-large} created the reference-free GEMBA-MQM metric, based on two versions of GPT, aimed at annotating MQM-based errors and evaluating the performance of their metric at system level using data from WMT'22 \citep{kocmi-etal-2022-findings} and WMT'23 \citep{kocmi-etal-2023-findings}. Their prompting is single-step and three-shot. They show that GEMBA-MQM achieves state-of-the-art performance compared with other metrics without human reference, and also outperforms many metrics with reference.

To the best of our knowledge, the latest work to date is that of \newcite{lu-etal-2025-mqm} with the MQM-APE metric, aiming to improve the quality of error annotations by 8 open-source LLMs with MQM without reference in order to boost the performance of MQM-APE over other baseline metrics at both system and segment level. They used data from WMT'22 \citep{kocmi-etal-2022-findings} for high-resource languages and IndicMT \citep{sai-b-etal-2023-indicmt} for low-resource languages. Their method consists of several steps: (a) MQM-based error annotation by LLMs using the GEMBA-MQM prompt; (b) post-editing by LLMs of annotated segments to determine errors that affect translation quality; (c) checking quality between pairs before and after post-editing to see whether PE improves the original translation. Errors that are not corrected are not counted as errors. The score of the original translation is calculated based on the errors counted after step (b). They then compare MQM-APE and GEMBA-MQM to show that MQM-APE improves performance at both the system and segment level, for high-resource and low-resource languages.

\paragraph*{LLMs for evaluating human translations} This overview shows the rapid development of this research field within the NLP community and for NLP purposes. However, a few works also focus on the use of LLMs, in particular ChatGPT, for practical purposes, including for translation training. For example, \newcite{araujo23comparing} used ChatGPT to evaluate translations by taking into account fluency, adequacy and appropriateness, each of these criteria being rated from 1 to 5 by ChatGPT. They compared ChatGPT annotations with reference annotations. The results show a consensus with regard to the lowest-scoring translation, but some variation in the best translations. Still, they show that ChatGPT is a reliable tool for researchers who regularly use MT to translate articles: ChatGPT can be useful for researchers who want to evaluate their machine translated texts, especially as it is an interactive tool offering recommendations, corrections, etc. 

\newcite{cao2023exploring} used ChatGPT in a pedagogical context. They compared 3 types of feedback for students (teacher feedback, self-feedback and ChatGPT feedback) on the basis of seven linguistic indicators (for lexicon, syntax and cohesion) and by evaluating the final versions after these feedbacks using the BLEU score \citep{papineni-etal-2002-bleu} with reference translations by professionals. They show that for cohesion and syntax, ChatGPT is no more useful than teacher feedback or self-feedback. On the other hand, ChatGPT improves students' lexis more than the other two types of feedback. They therefore suggest adopting a mixed approach for the three types of feedback, combining the capabilities of AI with the more conscious and nuanced feedback of teachers. One of our long-term goals is also to make use of the LLMs' capabilities in a pedagogical context, both by the students themselves and by teachers for evaluation purposes.

\section{Context and goals}

In this work, we are conducting experiments aimed at assessing the capabilities of LLMs to evaluate the quality of translations using prompting only, but with different motivations and objectives than those of the works outlined in Section~\ref{sec:rw}. We are investigating whether LLMs can identify and categorise errors in a translation, and particularly in specialised translation. In light of the works described in Section~\ref{sec:rw}, it seems appropriate to focus solely on prompting and not consider fine-tuning, since prompt-based evaluation already delivers strong results and, more importantly, the number of languages and domains for which error-annotated corpora of MT exist (especially for LSP translation) is too small for model fine-tuning to be considered a relevant solution.

In this context, given that our goal is to ask an LLM, namely ChatGPT, to identify and categorise errors, it is necessary to rely on an error typology that covers all the issues likely to arise in translations in order to ensure the effectiveness of this approach and the consistency of annotations. The typology we used is based on the MQM typology and on MeLLANGE (Multilingual e-learning in language engineering) \citep{castagnoli11designing,kubler08mellange}, an annotation framework designed for annotating translations in a translation training context. Even if it is not the main objective of this work, the possibility of using LLMs to identify errors in students' translations offers many interesting prospects, whether for evaluation assistance or to help students in their learning.

The modifications we have made\footnote{The full typology is described in Figure~\ref{fig:typology} in Appendix~\ref{sec:typology}.} make it possible to adapt these two typologies to the evaluation of specialised translation. This includes, among other factors, a more granular categorisation for terminological errors,\footnote{As shown in Figure~\ref{fig:typology}, the error category relating to terminology contains 10 error subtypes.} in order to account for the complex and domain-specific nature of such texts. 

\section{Experimental Method\label{sec:exp_method}}

\subsection{A prompt for identifying errors in translations \label{sec:prompt}}

The core of our work is based on the development of a prompt that enables an LLM to identify errors defined in a given typology. Unlike many research efforts in this field which, in line with the way translations are evaluated in the MT community, directly produce a score corresponding to the overall quality of a system or a translation, the prompt we have developed has a dual objective: to precisely locate the words and segments in the translation that are incorrect (the notion of "correctness" being defined by the error typology) and to characterise these errors by assigning them an error type (label) defined by the typology. After several trials and errors, we came up with a prompt whose results on a small set of examples seemed satisfactory enough to be systematically tested on a large scale\footnote{The prompt used was designed by a translator with prior experience in translation evaluation, but no extensive training in prompting and NLP, highlighting the fact that for this task and for the purposes at hand, it seems more appropriate to rely on an expert in translation evaluation rather than an expert in prompting, especially given the effectiveness of the prompt.}.

Our final prompt is a prompt in French\footnote{We carried out preliminary experiments with an English prompt, and the results indicated no significant difference between the English and the French prompts, although the latter performed slightly better on the sample tested.}, containing the instructions (task requested and its purpose, text type, explanation of attached file, expected output presentation), the error typology with a definition for each type of error, and the text to be annotated along with its source. In addition to the information contained in the instructions, we provide our full annotation manual\footnote{The annotation manual is a 50-page document designed to guide an evaluator in annotating translations according to our error typology. It provides general annotation guidelines, a full explanation of the typology, a definition and various examples for each error type.} as an attachment to the LLM. Given the large amount of text in the prompt, we used the prompt chaining technique\footnote{The prompt chaining technique involves linking multiple prompts together sequentially, where the output of one prompt becomes the input for the next one, enabling complex, multi-step reasoning or task completion.} \citep{ekin23prompt} and zero-shot mode, as no examples are included in the instructions. Although each text in the corpus was translated at the document level, and not by sentence, we opted for sentence-level alignment when using ChatGPT to annotate errors, in order to minimise the volume of densely-packed information to be processed by the model. The full prompt is given in Figure~\ref{fig:prompt} (Appendix~\ref{sec:full_prompts}).

Note that, with the exception of one sentence specifying the type of text translated (abstracts of research articles in NLP), our prompt does not contain any instructions specifically relating to the text type or to the (highly) specialised domain. Therefore, although we have not specifically tested this aspect, it seems likely that the results we report in this work can also be applied to other types of text.

In our experiments, we experiment with two variations of this prompt and use it to identify errors in translations produced by different mainstream translation systems.

\subsection{Reference human annotations}

To evaluate the ability of an LLM to identify errors in the translation of a specialised text, we built a corpus comprising source documents (abstracts of NLP research articles in English from the HAL open archive\footnote{\url{https://hal.science/}}), their translation in French by different MT systems used by both the general public and professional translators (namely DeepL and ChatGPT\footnote{Here is the prompt (translated in English) we used to translate the texts with ChatGPT: “You are a translator who specialises in translating research articles on natural language processing. Translate the following text into French, respecting the structure of the original text and not omitting any elements.”}) and an annotation of these translations by a human expert (a professional translator with extensive experience in evaluating translations and using our error typology) who identified the errors contained in these MT outputs. In this context, "annotation" refers to the manual identification and labelling of errors in a translation, where the annotator identifies incorrect segments and assigns one or more error categories based on our predefined typology. The annotated translations contain error spans and error type labels (occasionally several possible labels) for each error.

In the end, our corpus\footnote{Our corpus of French translation reference annotations with English source texts is available here: \url{https://doi.org/10.34847/NKL.52E571A3}} is divided into two sub-corpora: a) a sub-corpus of 35 source texts translated by DeepL with the annotated translations based on the error typology (10,500 words\footnote{To count the number of words, we naively tokenised our corpus using spaces.}), and b) a sub-corpus of 25 source texts translated by ChatGPT with the annotations based on the typology (7,431 words). 

Figure~\ref{fig:ex_annot} shows an example of the annotation produced by our expert. In the first sub-corpus, the expert identified 399 errors (an average of 11.4 per document); the errors ranged from 2 to 81 characters (average: 15 characters), and had between 1 and 6 possible labels (average: 2.3). For the second sub-corpus, the expert identified 193 errors (on average 7.7 per document); the errors ranged from 2 characters to 103 characters (average: 22 characters), and had between 1 and 4 possible labels (average: 2.1).

\begin{figure*}
\centering
\fbox{
\begin{minipage}{.9\textwidth}
Les contes de fées, les \err{contes du peuple}\textsubscript{LA-TL-INS, LA-TL-ING} et plus généralement les \err{histoires d’enfants}\textsubscript{TR-DI, LA-SY-PR, LA-SY-GNC, LA-TL-INS, LA-TL-ING} ont récemment attiré la communauté du Traitement Automatique des Langues (TAL). \err{A ce titre}\textsubscript{LA-HY-PU} très peu de corpus existent, et les ressources linguistiques manquent. Le travail présenté dans cet article vise à combler \err{la lacune}\textsubscript{LA-UR, LA-TC-CE, LA-TC-CN, LA-SY-DET, LA-ST-AW} en présentant un corpus annoté syntaxiquement et sémantiquement. \err{Elle}\textsubscript{LA-IA-GE, LA-UR, LA-TC-CE, LA-TC-CN} se \err{focusse}\textsubscript{TR-SI-UT, TR-SI-TL, LA-TL-ING} sur l'analyse linguistique d'un corpus de contes de fées et fournit une description des ressources syntaxiques et sémantiques développées pour \err{l'extraction des informations}\textsubscript{LA-TL-INS, LA-SY-DET, LA-SY-PR}.
\end{minipage}
}
\caption{Example of a human reference annotation: each error is identified by its span (text written on an orange background) and one or more labels (in subscript). \label{fig:ex_annot}}
\end{figure*}

\subsection{Evaluating the evaluations \label{sec:eval}}

In order to automatically evaluate the performance of our prompts in detecting errors identified by the expert translator, we use the standard recall and precision metrics commonly employed in NLP to assess error detection systems. Precision measures the proportion of errors identified by an LLM with our prompt that are actually correct. It is calculated as the ratio of true positives (correct corrections) to the total number of corrections made (true positives + false positives). Precision reflects the system's ability to avoid making incorrect corrections. Conversely, recall gauges our prompts's capacity to pinpoint all errors present in a text. It is calculated as the ratio of true positives to the total number of actual errors in the corpus. This number is given by the sum of the number of true positives (the number of errors correctly identified) and of false negatives (the number of errors “missed” by the model).

A high precision indicates that our system makes very few incorrect corrections, but it does not necessarily mean that all existing errors are detected. Conversely, a high recall shows that the system identifies most of the errors in a text but might introduce many false corrections, leading to lower precision. To easily compare the performance of the different prompts we consider, we use the standard F$_1$ score, which combines recall and precision into a single number to compare the performance of two systems.

Defining these three metrics involves determining whether an error identified by the expert corresponds to a predicted error. However, this is not always straightforward, as the definition of an error can be subjective and open to interpretation: the decision of whether to include a word in the definition of an error can vary between annotators. For practical reasons, we decided to consider an error in the reference annotation as correctly identified by the LLM if the error shares at least one character with a predicted error.\footnote{We have also ensured in our evaluation that a reference error is not associated with two different predicted errors.} This decision is based on the assumption that even a single shared character is enough to draw a translator's attention to the area with a potential issue.

With these definitions, precision $P$ and recall $R$ are simply defined as : 
\begin{equation}
P = \frac{\textrm{number of errors correctly identified}}{\textrm{number of predicted errors}}
\end{equation}
and
\begin{equation}
R = \frac{\textrm{number of errors correctly identified}}{\textrm{number of errors in reference}}
\end{equation}

In our evaluation, precision and recall will be calculated at the level of each document, enabling a fine-grained analysis of the system's performance across individual texts. The results will then be averaged over all documents, meaning the reported numbers correspond to macro-recall and macro-precision.

In several cases, an error can be tagged with more than one error label; this is the case, for example, with terminological errors, which can distort the meaning of the message (terminological error + content transfer error). In order to assess a prompt's ability to correctly categorise errors, we also report, for each experiment, the proportion of correctly identified errors whose predicted label matched at least one reference label, since the model predicts only one label for each error. Asking the LLM to predict multiple labels would make the task too complex, both for the LLM and our meta-evaluation.

\subsection{Experiments}

So far, three different experiments have been carried out. The first experiment, denoted “\texttt{long prompt}” in the following, consisted in having 35 MT outputs in French from DeepL evaluated by ChatPT \citep{openai2024gpt4technicalreport}\footnote{We used the version of ChatGPT that relies on GPT-4o.} with the prompt described in Section~\ref{sec:prompt}. The second, denoted “\texttt{short prompt}”, with the same 35 MT outputs, involved testing a shorter and less information-laden version of the prompt, i.e.\ removing the definitions of each type of error; this follows suggestions made by \newcite{lu-etal-2024-error}, who recommend against providing error descriptions in detail. Finally, the last experiment involved ChatGPT evaluating 25 MT outputs of other source texts it had generated itself.

The primary aim of these experiments is, firstly, to see whether ChatGPT can perform annotation tasks on specialised translations with our error typology. Next, we aim to understand the strengths and weaknesses of this model, especially in terms of error identification and categorisation. We also intend to see whether defining each error in the prompt influences the quality of annotations and whether its capabilities vary according to the source of the MT output (DeepL or its own translations).

\section{Results\label{sec:results}}

\begin{table}
    \centering
    {\footnotesize
    \begin{tabular}{llccc}
\toprule
&& \multicolumn{3}{c}{MT System} \\
\cline{3-5}
&& DeepL & & ChatGPT \\
\midrule
&\# texts               &  35 && 25\\
&\# gold errors         & 399 && 193 \\
\midrule
\multicolumn{5}{l}{\textit{long prompt}} \\
& \# pred. errors      & 384                        && 224 \\
& precision            & 0.792 {\tiny $\pm$ 0.0396} && 0.47 {\tiny $\pm$ 0.0989 } \\
& recall               & 0.653 {\tiny $\pm$ 0.0488} && 0.57 {\tiny $\pm$ 0.107 }\\
& F$_1$                & 0.707 {\tiny $\pm$ 0.0393} && 0.496 {\tiny $\pm$ 0.0933} \\
& \% correctly labeled & 64.1\,\% && 45.3\,\% \\
\midrule
\multicolumn{5}{l}{\textit{short prompt}} \\
& \# pred. errors     & 417                        && --- \\
& precision           & 0.745 {\tiny $\pm$ 0.0575} && ---   \\
& recall              & 0.671 {\tiny $\pm$ 0.0505} && --- \\
& F$_1$               & 0.702 {\tiny $\pm$ 0.0531} && --- \\
& \% correctly labeled  & 46.9\,\% && --- \\
\bottomrule
\end{tabular}}
\caption{Results achieved by our different prompts on the two corpora we consider. “\# gold errors” represents the number of errors found by the expert annotator, “\#pred error” represents the number of errors predicted by our system. We have computed the 95\% confidence intervals for the different scores we consider using the bca bootstrap method of \newcite{effron93introduction}. \label{tab:res}}
\end{table}

\begin{figure*}
\centering
\includegraphics[width=.9\textwidth]{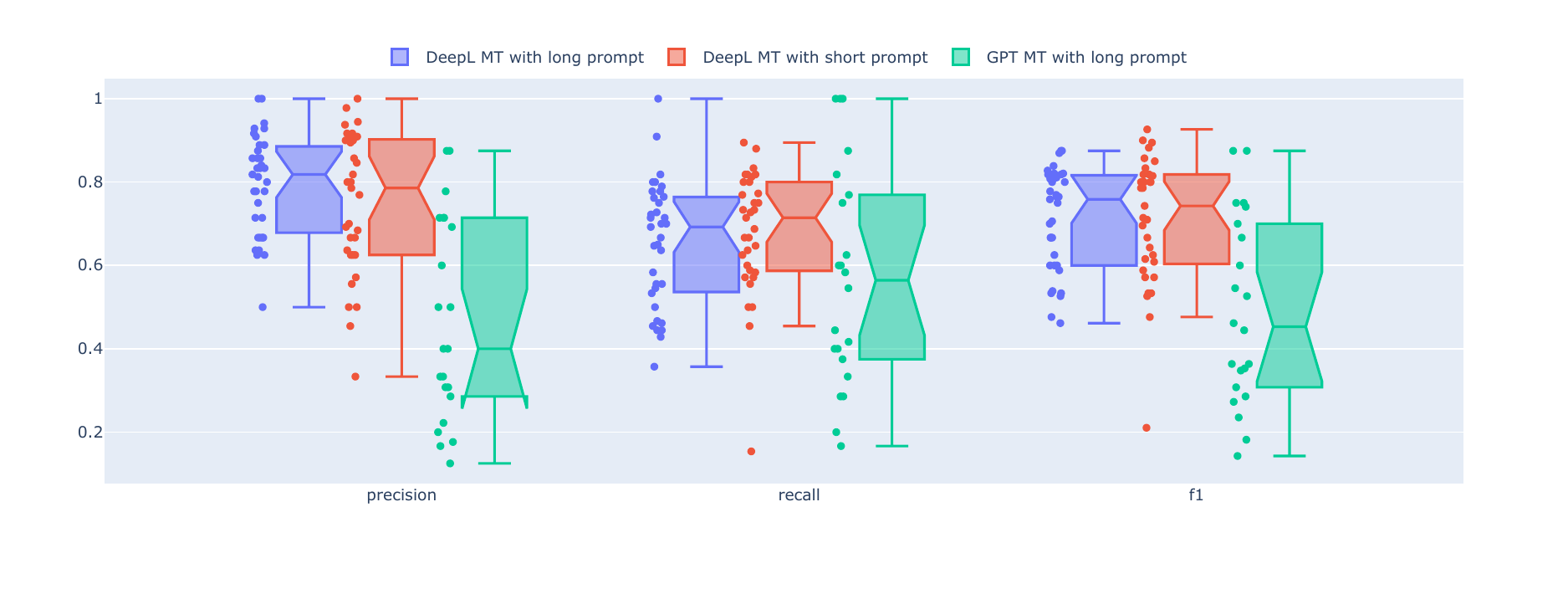}
\caption{Distribution of precisions, recalls and F$_1$ scores across documents for the different prompts we consider. \label{fig:distrib}}
\end{figure*}

As explained in Section~\ref{sec:eval}, we measure the ability of the different prompts considered to correctly identify translation errors by evaluating recall, precision and F$_1$ score on two corpora of translations generated by different MT systems. The results obtained are summarised in Table~\ref{tab:res}. For the sake of clarity, we have also reported in this Table the total number of gold errors (i.e.\, errors identified by an expert) in our corpus, the number of predicted errors and the percentage of labels that are correctly identified. Figure~\ref{fig:distrib} also shows the distribution of the various scores obtained to enable a more detailed analysis of the performance of the prompts. In the remainder of this Section, we will detail the results achieved for each experiment.

\paragraph*{Annotations of DeepL MTs}

Using ChatGPT with the \emph{long} prompt to identify and categorise errors in DeepL MT outputs shows promising results.

For all errors identified in reference human annotations, ChatGPT identifies between 6 and 7 out of 10. The model also seems to perform very well when it comes to categorising errors based on the error typology, managing to accurately categorise around 65\% of them. If a sentence contains no errors ---~which does happen~---, ChatGPT occasionally acknowledges the fact that there are no errors in the sentence, but it can also over-annotate the translation by detecting errors that are not actually errors (what we call “false errors” here). On average, in an annotated text, ChatGPT identifies between 1 and 2 “false” errors, i.e.\ errors that are not identified as errors in the reference annotation. The number of false errors varies between 0 and 5 per text. These false errors represent 14.47\% of the errors annotated by the model.

Although the average F$_1$ score (0.71) indicates a satisfactory overall performance, the dispersion of scores (Figure~\ref{fig:distrib}) shows that the model can react unpredictably to different texts: depending on the document, precision may vary from $1.0$ to $0.5$ and recall from $1.0$ to $0.35$. This variability could reflect sensitivity to differences in the complexity or nature of the errors to be identified, making performance occasionally more random depending on the case. These rather unpredictable performances of ChatGPT have already been pointed out by the scientific community (see, for example, \newcite{siu2023ChatGPT}). However, it does call into question the practical interest of the model: it is unlikely that a translator would use such a system to identify errors if they were randomly wrong.

Using a shorter prompt by removing the definition given to each type of error in the instructions (see \textsection~\ref{sec:prompt}) shows similar results. The system's ability to identify errors is more or less the same: the overall F$_1$ score is also around 0.70. Whereas, with the full prompt, ChatGPT performed better in error categorisation than in error identification, the opposite happens with the short prompt. In fact, it identifies almost 7 out of 10 errors. On the other hand, around 5 out of 10 errors are incorrectly categorised.

This drop in error categorisation performance comes as no surprise, since the prompt no longer contains the definitions of each type of error. However, it is more surprising to see that the removal of this information has only a (very) slight impact on the system's ability to identify errors, suggesting that ChatGPT's ability to identify translation errors is not linked to the information it has extracted from the prompt, but only to the knowledge it has acquired during its training or to the knowledge it acquires from the attached annotation manual.

As far as false errors are concerned, the average here is 1.71, and per text, the number of false errors varies between 0 and 7. False errors account for 17.86\% of all errors annotated by ChatGPT with the short prompt. For this test, recall is slightly higher than in the first experiment (0.67 compared with 0.65), and we observe a slight loss of precision, reaching 0.75.

It is also interesting to note that, as shown in Figure~\ref{fig:distrib}, removing the definition for each error type from the prompt significantly increases performance variability: In contrast to the performance of the long prompt, where the lowest precision was 0.500, here several texts (6) show clearly low scores, highlighting specific difficulties or cases where the model performs less well. This comparison highlights a more marked uncertainty in the reliability of the model's evaluations on this set of texts with the short prompt.

\paragraph*{Annotations of ChatGPT MT outputs}

ChatGPT annotations of its own 25 MT outputs with the full prompt show particularly weak results compared to the two previous experiments.

Table~\ref{tab:res} shows that when annotating its own MT outputs, ChatGPT identifies only about half of the errors contained in the reference annotations. Well-categorised errors are also below 50\%. The rate of false errors per text doubles or even triples compared with the two previous experiments, reaching almost 5 false errors per annotated text. They account for 55.02\% of all errors identified by the model, i.e.\ more than half, and range from 0 to 14 per text.

The average overall F$_1$ score is significantly lower than it was in the two previous experiments, dropping to 0.496. In terms of recall and precision, the results are no better, with a recall of 0.57 and a precision of 0.47.

Figure~\ref{fig:distrib} shows a low average score and high variability, reflecting limited performance in this particular setting. This can be explained by the fact that ChatGPT evaluated its own machine translated texts, a task that seems to raise specific challenges. The large number of scores below 0.5 suggests that the model struggles to identify and categorise its own errors in a systematic way, probably due to implicit bias or a lesser ability to step back from its own productions. This contrast with the other evaluation scenarios highlights a weakness in the model's self-evaluation.

\section{Discussion}

Results achieved with the full prompt for DeepL MT annotations show satisfactory performances of ChatGPT, with an average F$_1$ score of 0.707 and a fairly strong capability to identify and categorise errors (about 65\%). In contrast to what has been claimed by the community (see e.g.\ \newcite{lu2023error}), our experiments show that providing a detailed definition for each type of error in the prompt slightly improves the model's performance, in particular for error categorisation, which was below 50\% with the short prompt and reached 65\% with the definitions in the prompt.

Despite these encouraging results, however, the variability in scores, particularly with the short prompt and with ChatGPT's MT outputs, suggests that ChatGPT's effectiveness with these parameters is not consistent. Indeed, a large number of texts in these two experiments have very low recall, precision and F$_1$ scores.

Finally, ChatGPT shows considerable limitations in self-assessment, with significantly lower performance when annotating its own translations (F$_1$ score of 0.496). This result highlights a possible bias and the model's lack of ability to objectively evaluate its own productions, which justifies our motivations to set up annotation experiments with other open-source LLMs, as have \newcite{fernandes-etal-2023-devil} and \newcite{lu-etal-2025-mqm}, demonstrating that comparable ---~or even better~--- performance can also be achieved with open-source LLMs.

\begin{figure*}
    \centering
    \includegraphics[width=0.5\linewidth]{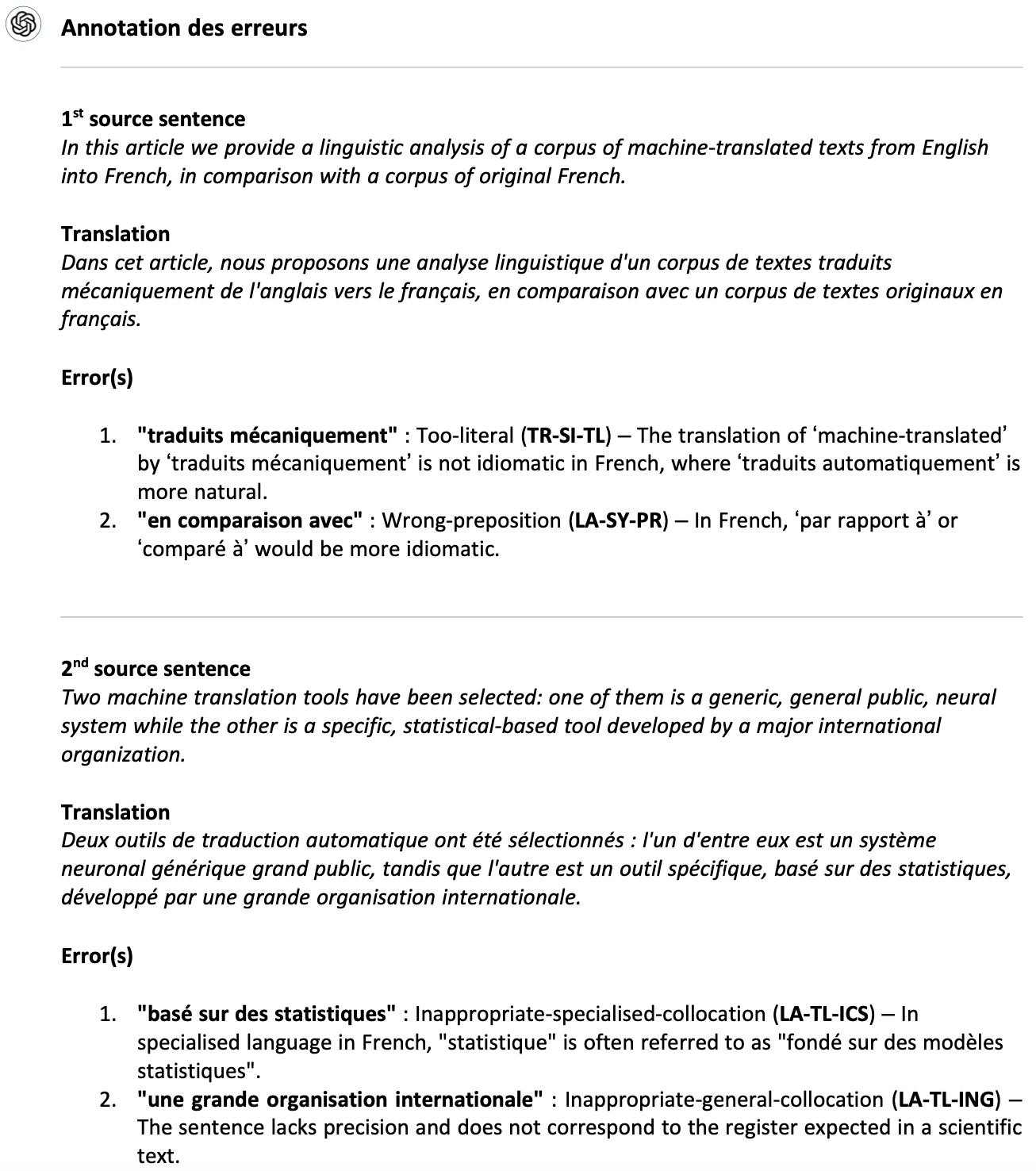}
    \caption{Example of annotation by ChatGPT. This figure shows the sentence-by-sentence annotation performed by ChatGPT, which identifies the error, categorises it, assigns a label and provides explanations and solutions for improvement. The initial output of ChatGPT is in French, since the prompt provided is written in French. For the purposes of this article, we have translated it into English.}
    \label{fig:exemple-anno2}
\end{figure*}

In order to further assess the relevance and usability of ChatGPT's outputs, we provide an example of an individual annotation\footnote{Figure~\ref{fig:exemple-anno2} represents an annotation produced by ChatGPT. However, this is not the output we considered for calculating comparison scores with the reference annotations. To calculate these scores, we asked ChatGPT to convert its annotations into a table that could be used and analysed automatically.} performed by the model (see Figure~\ref{fig:exemple-anno2}). This annotation by ChatGPT clearly shows that the LLM annotates the errors as instructed in the prompt, respecting the requested output format (sentence by sentence), giving the error span, the error category, the associated label and an explanation. However, we did not directly request explanations for each annotated error in our prompt, but the model seems to do so systematically. For these sentences, the explanations the model provides make sense. These explanations highlight the potential usability of these experiments with ChatGPT: since the LLM provides explanations and possible solutions for each potential error, it seems reasonable to consider conducting other experiments in a real-life classroom setting using ChatGPT's annotations (see Section~\ref{sec:future}).

\section{Conclusion\label{sec:conclusion}}

This study explored the use of ChatGPT for annotating MT outputs based on a customised error typology adapted to our specific needs in a specialised translation training setting. The annotations generated by the model were compared with reference human annotations to evaluate its ability to identify and categorise errors in a translation generated by DeepL or ChatGPT. Initial results are encouraging, particularly with external machine translations, where ChatGPT identified and categorised most errors with reasonable accuracy, in particular with the long prompt containing the definition for each type of error. However, its performance was far less reliable when evaluating its own translations. 

Another key finding from our experiments is that, given the lack of a significant difference in error identification between the full and short prompts, it seems reasonable to suggest that the structure and degree of detail of the prompt does not have a major impact on ChatGPT's performance in this annotation task. This could indicate that ChatGPT is performing its annotations efficiently even without detailed instructions (without a definition for each error), relying more on its knowledge acquired during training rather than on the specifications of the prompt. However, this also suggests that while LLMs can rely on their pre-trained knowledge to identify errors, their ability to categorise these errors correctly benefits from clear, structured instructions and definitions.

\paragraph*{Future Work\label{sec:future}} Future experiments will extend this research to open-source LLMs, focusing on their potential to provide annotations of comparable or superior quality. These models, with greater transparency, will be evaluated not only for their accuracy and capabilities in annotating translations but also for their ease of integration into automated workflows for translation quality assessment.

Ultimately, our aim is to test the effectiveness of this automated evaluation by LLMs in a practical context of translation training. Firstly, we intend to optimise the human evaluation process. Specifically, with teachers annotating students' translations, we aim to examine whether the use of annotations generated by LLMs can reduce the cognitive effort associated with the annotation process. Additionally, we intend to carry out experiments with translation students and test whether the use of LLM annotations help them improve the quality of their MT post-editing. Our aim is to test our prompt with other domains, notably earth and planetary science.

\section*{Limitations}

ChatGPT, as an OpenAI proprietary model, has some limitations that need to be taken into account in our experiments. The lack of transparency regarding its training data, the uncertainties associated with its availability in the future and the fluctuations in its performance over time make it difficult to assess its capabilities in a rigorous and reproducible way. These issues have been highlighted by other researchers, notably \newcite{chen24how}, who observed significant variations over the course of 2023. That being said, ChatGPT remains a mainstream tool that is used by many translators in their day-to-day work. Therefore, we believe that it would be relevant to evaluate it.  However, we intend to conduct similar experiments with other open-source LLMs, which have already demonstrated state-of-the-art performance. These models offer greater transparency and full control over the versions used, which is essential to guarantee traceable and reproducible results.

\section*{Acknowledgments}
This research was funded by the French Agence Nationale de la Recherche (ANR) under the project MaTOS - ``ANR-22-CE23-0033-03''.

\bibliography{anthology,mtsummit25}

%\bibliography{mtsummit25}

\appendix

\section{Error typology \label{sec:typology}}

\begin{figure*}
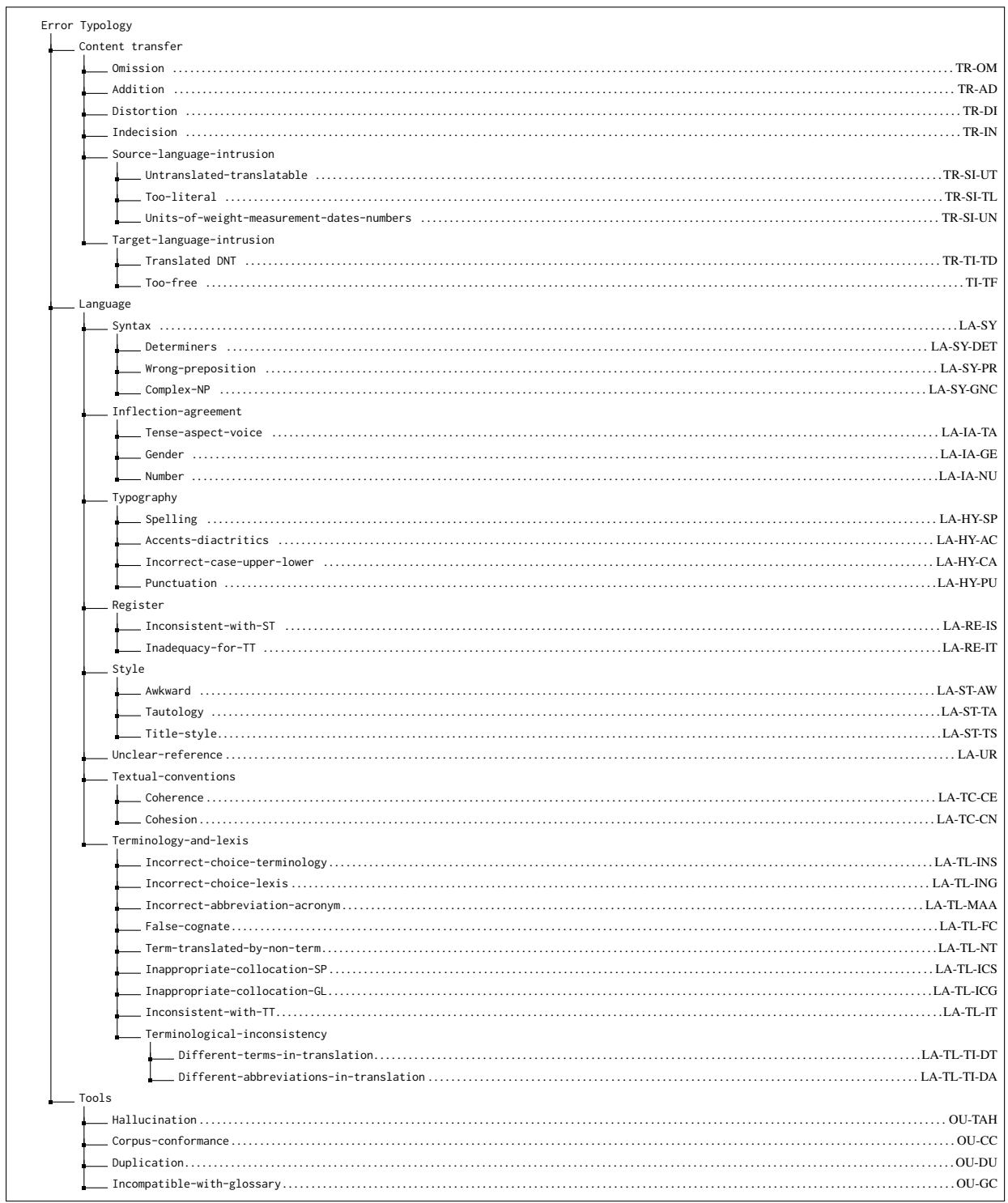

\fbox{\begin{minipage}{\textwidth}
\renewcommand*\DTstyle{\ttfamily\tiny}
\setlength{\DTbaselineskip}{10pt}
\dirtree{%
.1 Error Typology.
.2 Content transfer.
.3 Omission   \DTcomment{TR-OM}.
.3 Addition   \DTcomment{TR-AD}.
.3 Distortion \DTcomment{TR-DI}.
.3 Indecision \DTcomment{TR-IN}.
.3 Source-language-intrusion.
.4 Untranslated-translatable \DTcomment{TR-SI-UT}.
.4 Too-literal \DTcomment{TR-SI-TL}.
.4 Units-of-weight-measurement-dates-numbers \DTcomment{TR-SI-UN}.
.3 Target-language-intrusion.
.4 Translated DNT \DTcomment{TR-TI-TD}.
.4 Too-free \DTcomment{TI-TF}.
.2 Language.
.3	Syntax \DTcomment{LA-SY}.
.4 Determiners \DTcomment{LA-SY-DET}.
.4 Wrong-preposition \DTcomment{LA-SY-PR}.
.4 Complex-NP \DTcomment{LA-SY-GNC}.
.3 Inflection-agreement.
.4 Tense-aspect-voice \DTcomment{LA-IA-TA}.
.4 Gender \DTcomment{LA-IA-GE}.
.4 Number \DTcomment{LA-IA-NU}.
.3 Typography.
.4 Spelling \DTcomment{LA-HY-SP}.
.4 Accents-diactritics \DTcomment{LA-HY-AC}.
.4 Incorrect-case-upper-lower \DTcomment{LA-HY-CA}.
.4 Punctuation \DTcomment{LA-HY-PU}.
.3 Register.
.4 Inconsistent-with-ST \DTcomment{LA-RE-IS}.
.4 Inadequacy-for-TT \DTcomment{LA-RE-IT}.	
.3 Style.
.4 Awkward \DTcomment{LA-ST-AW}.
.4 Tautology \DTcomment{LA-ST-TA}.
.4 Title-style\DTcomment{LA-ST-TS}.
.3 Unclear-reference\DTcomment{LA-UR}.
.3 Textual-conventions.
.4 Coherence\DTcomment{LA-TC-CE}.
.4 Cohesion\DTcomment{LA-TC-CN}.
.3 Terminology-and-lexis.
.4 Incorrect-choice-terminology\DTcomment{LA-TL-INS}.
.4 Incorrect-choice-lexis\DTcomment{LA-TL-ING}.
.4 Incorrect-abbreviation-acronym\DTcomment{LA-TL-MAA}.
.4 False-cognate\DTcomment{LA-TL-FC}.
.4 Term-translated-by-non-term\DTcomment{LA-TL-NT}.
.4 Inappropriate-collocation-SP\DTcomment{LA-TL-ICS}.
.4 Inappropriate-collocation-GL\DTcomment{LA-TL-ICG}.
.4 Inconsistent-with-TT\DTcomment{LA-TL-IT}.
.4 Terminological-inconsistency.
.5 Different-terms-in-translation\DTcomment{LA-TL-TI-DT}.
.5 Different-abbreviations-in-translation\DTcomment{LA-TL-TI-DA}.
.2 Tools.
.3 Hallucination\DTcomment{OU-TAH}.
.3 Corpus-conformance\DTcomment{OU-CC}.
.3 Duplication\DTcomment{OU-DU}.
.3 Incompatible-with-glossary\DTcomment{OU-GC}.
}
\end{minipage}}
\caption{The error typology used in our experiments. \label{fig:typology}}
\end{figure*}

\section{Full prompts \label{sec:full_prompts}}
\begin{figure*}
\small
%\begin{tabularx}{\textwidth}{cX}
%  \toprule
\begin{enumerate}
\item \begin{Verbatim}[breaklines=true]
Tâche : annoter une traduction 
Objectif : repérer des erreurs sur la base d’une typologie d’erreurs que je te fournis. 
Type de texte : résumé d’article scientifique dans le domaine du TAL
Fichier joint : MANUEL D’ANNOTATION, qui contient des explications plus détaillées et des exemples des types d’erreurs que je vais te fournir ci-dessous. 
Présentation de la sortie :
- 1re phrase source
- 1re phrase cible dans la traduction
- liste les erreurs
Etc. jusqu’à la fin de la traduction
---------------
Je vais te donner la typologie d’erreurs.
\end{Verbatim}
\item \begin{Verbatim}[breaklines=true]
Typologie d’erreurs à suivre méticuleusement : veille à utiliser les types d’erreurs présents et n’en invente aucun. De même, respecte les codes liés à chaque type d’erreur à la lettre ; ne prends donc aucune liberté. 
Explication de la typologie : elle est divisée en 3 grandes catégories d’erreurs : les erreurs de transfert de contenu (erreurs altérant le sens du message ou entravant sa compréhension), les erreurs de langue, et les erreurs liées aux outils ou à leur maîtrise. 
Voici la typologie :
1.	Transfert-contenu (GRANDE CATÉGORIE, NE PAS UTILISER)	
1.1.	Omission_TR-OM
* Une omission se produit lorsqu’il manque, dans la traduction, une idée qui est présente dans le texte source. Il ne faut pas confondre omission et implicitation. Une omission a lieu sans réelle raison valable, alors qu’une implicitation est un moyen d’éviter une surtraduction.	
1.2.	Rajout_TR-AD	
* À l’instar de la différence entre omission et implicitation, on peut souligner une différence de nuance entre le rajout et l’explicitation. L’ajout est considéré comme une erreur, alors que l’explicitation peut s’expliquer par le fait que le traducteur ou le post-éditeur souhaite éviter la sous-traduction.
... jusqu’au bout de la typologie ...
----------- 
- Prête attention à tous les aspects, autant le transfert de contenu que la langue et la terminologie et les erreurs liées aux outils. 
- Si tu as besoin d’exemples, réfère toi au manuel d’annotation en pièce jointe.
-----------
Je vais te donner la traduction à évaluer avec son texte source.
\end{Verbatim}
\item \begin{Verbatim}[breaklines=true]
  Voici le texte source et sa traduction à annoter :
  (source text)
  (target text)
  ----------
  PROCÈDE À L’ANNOTATION. Attention, n’annote QUE les erreurs, pas des améliorations ou suggestions ! Il peut y avoir plusieurs erreurs dans une même phrase.  
\end{Verbatim}
\end{enumerate}
\caption{Prompt used on GPT-4o \label{fig:prompt}}
\end{figure*}

%\section{Example Appendix}
%\label{sec:appendix}

%This is an appendix.

\end{document}